\documentclass[10pt,twocolumn,letterpaper]{article}

\usepackage{cvpr}
\usepackage{times}
\usepackage{epsfig}
\usepackage{graphicx}
\usepackage{amsmath}
\usepackage{amssymb}
\usepackage{bm}
\usepackage{lipsum}
\usepackage{color}
\usepackage{booktabs}
\usepackage{algorithm}

\usepackage[pagebackref=true,breaklinks=true,letterpaper=true,colorlinks,bookmarks=false]{hyperref}

\cvprfinalcopy 

\ifdefined\ShowNotes
  \newcommand{\colornote}[3]{{\color{#1}\bf{#2: #3}\normalfont}}
\else
  \newcommand{\colornote}[3]{}
\fi

\newcommand {\amir}[1]{\colornote{blue}{Amir}{#1}}

\def\cvprPaperID{2567} 

\ifcvprfinal\fi
\begin{document}

\makeatletter
\newcommand{\raisemath}[1]{\mathpalette{\raisem@th{#1}}}
\newcommand{\raisem@th}[3]{\raisebox{#1}{$#2#3$}}
\makeatother

\title{Graph Embedded Pose Clustering for Anomaly Detection}


\author{
Amir Markovitz\textsuperscript{1}, Gilad Sharir\textsuperscript{2}, Itamar Friedman\textsuperscript{2}, Lihi Zelnik-Manor\textsuperscript{2}, and Shai Avidan\textsuperscript{1} \\ \\
\textsuperscript{1}Tel-Aviv University, Tel-Aviv, Israel \qquad \textsuperscript{2}Alibaba Group 
\\ {\tt\small \{markovitz2@mail, avidan@eng\}.tau.ac.il, \{first.last\}@alibaba-inc.com}
}
\maketitle

\begin{abstract}
We propose a new method for anomaly detection of human actions. Our method works directly on human pose graphs that can be computed from an input video sequence. This makes the analysis independent of nuisance parameters such as viewpoint or illumination. We map these graphs to a latent space and cluster them. Each action is then represented by its soft-assignment to each of the clusters. This gives a kind of "bag of words" representation to the data, where every action is represented by its similarity to a group of base action-words. Then, we use a Dirichlet process based mixture, that is 
useful for handling proportional data such as our soft-assignment vectors, to determine if an action is normal or not.

We evaluate our method on two types of data sets. The first is a fine-grained anomaly detection data set  (e.g. ShanghaiTech) where we wish to detect unusual variations of some action. The second is a coarse-grained anomaly detection data set (e.g.,\ a Kinetics-based data set) where few actions are considered normal, and every other action should be considered abnormal.

Extensive experiments on the benchmarks show that our method\footnotemark performs considerably better than other state of the art methods.
\end{abstract}
\footnotetext{Code available at: \url{https://github.com/amirmk89/gepc}
}

\section{Introduction}



Anomaly detection in video has been investigated extensively over the years. This is because the amount of video captured far surpasses our ability to manually analyze it. Anomaly detection algorithms are designed to help human operators deal with this problem. The question is how to define anomalies and how to detect them.

The decision of whether an action is normal or not is nuanced. In some cases, we are interested in detecting abnormal variations of an action. For example, an abnormal type of walking. We term this fine-grained anomaly detection. In other cases, we might be interested in defining normal actions and regard any other action as abnormal. For example, we might be interested in determining that dancing is normal, while gymnastics are abnormal. We call this coarse-grained anomaly detection.


We desire an algorithm that can handle both types of anomaly detection in a single, unified fashion. Such an algorithm should take as input an unlabeled set of videos that capture normal actions \emph{only} (fine- or coarse-grained) and use that to train a model that will distinguish normal from abnormal actions.






We take advantage of the recent progress in human pose estimation and assume our algorithm takes human pose graphs as input. This offers several advantages. First, it abstracts the problem and lets the algorithm focus on human pose and not on irrelevant features such as viewing direction, illumination, or background clutter. In addition, a human pose can be represented as a compact graph, which makes analyzing, training and testing much faster.

\begin{figure*} [!h]	
\hspace*{-0.0cm}\begin{tabular}{c}
\includegraphics[width=\textwidth]{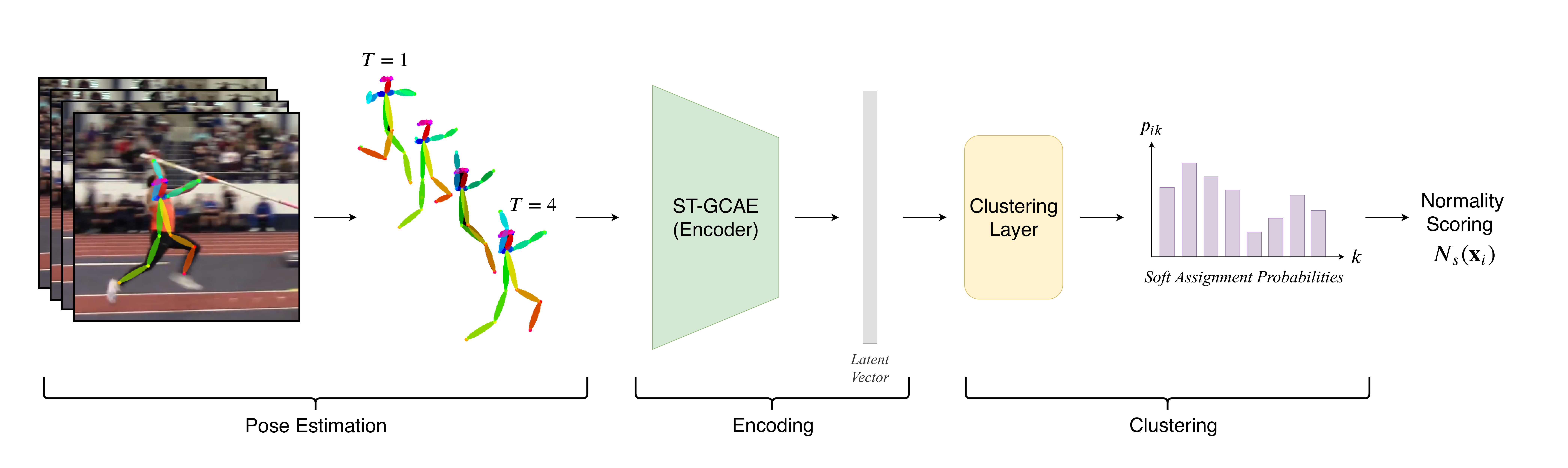} \vspace{0.0cm}
\end{tabular}
\caption{{\bf Model Diagram (Inference Time):} To score a video, we first perform pose estimation. The extracted poses are encoded using the encoder part of a Spatio-temporal graph autoencoder (ST-GCAE), resulting in a latent vector. The latent vector is soft-assigned to clusters using a deep clustering layer, with ${p_{ik}}$ denoting the probability of sample~$\mathbf{x}_i$ being assigned to cluster~$k$.}

\vspace{-0.2cm}
\label{figure:model_diagram}
\end{figure*}

Given a sequence of video frames, 
we use a pose estimation method to extract the keypoints of every person in each frame. 
Every person in a clip is represented as a temporal pose graph. We use a combination of an autoencoder and a clustering branch to map the training samples into a latent space where samples are soft clustered. 
Each sample is then represented by its soft-assignment to each of the $k$ clusters. This can be understood as learning a bag-of-words representation for actions. Each cluster corresponds to an action-word, and each action is represented by its similarity to each of the action-words.
Figure~\ref{figure:model_diagram} gives an overview of our method. 


The soft-assignment vectors capture proportional data and the tool to measure their distribution is the Dirichlet Process Mixture Model. Once we fit the model to the data, we can obtain a normality score for each sample and determine if the action is to be classified as normal or not.



The algorithm thus consists of a series of abstractions. Using human pose graphs eliminates the need to deal with viewpoint and illumination changes. And the soft-assignment representation abstracts the type of data (fine-grained or coarse-grained) from the Dirichlet model.

We evaluate our algorithm in two settings. The first is the \textit{ShanghaiTech Campus}~\cite{luo2017revisit} dataset, a large and extensively evaluated anomaly detection benchmark. 
This is a typical (\emph{fine-grained}) anomaly detection benchmark in which normal behavior is taken to be walking, and the goal is to detect abnormal events, such as people running, fighting, riding bicycles, throwing objects, etc.

The second is a new problem setting we propose, and denote \emph{Coarse-grained} anomaly detection. Instead of focusing on a single action (i.e., walking), as in the ShanghaiTech dataset, we construct a training set consisting of a varying number of actions that are to be regarded as normal. 
For example, the training set may consist of video clips of different dancing styles. At test time, every dance video should be classified as normal, while any other action should be classified as abnormal.

We demonstrate this new, challenging, \emph{Coarse-grained anomaly detection} setting on two action classification datasets. First is the \textit{NTU-RGB+D} dataset, where 3D body joints are detected using Kinect. Second is a larger and more challenging dataset that consists of 250 out of the 400 actions in the \textit{Kinetics400} dataset\footnote{We only use a subset of the classes as not all classes can be detected using human pose detectors.}. For both datasets, we use a subset of the actions to define a training set of normal actions and use the rest of the videos to test if the algorithm can correctly distinguish normal from abnormal videos.


We conduct extensive experiments, compare to a number of competing approaches and find that our algorithm outperforms all of them.

\noindent
To summarize, we propose three key contributions: \vspace{-0.25cm}
\begin{itemize}
\itemsep-0.5em 

\item The use of embedded pose graphs and a Dirichlet process mixture for video anomaly detection;
\item A new \emph{coarse-grained} setting for exploring broader aspects of video anomaly detection;
\item State-of-the-art AUC of $0.761$ for the \emph{ShanghaiTech Campus} anomaly detection benchmark.
\end{itemize}

\section{Background}

\subsection{Video Anomaly Detection}
The field of anomaly detection is broad and has a large variation in setting and assumptions, as is evident by the different datasets  proposed to evaluate methods in the field. 

\amir{removed old datasets paragraph}
For our fine-grained experiment, we use the \textit{ShanghaiTech Campus dataset}~\cite{luo2017revisit}. Containing 130 anomalous events in 13 different scenes, with various camera angles and lighting conditions, it is more diverse and significantly larger than all previous common  datasets. It is presented in detail in section~\ref{stc_dataset}.

In recent years, numerous works tackled the problem of anomaly detection in video using deep learning based models. Those could be roughly categorized into reconstructive models, predictive models, and generative models. 

Reconstructive models learn a feature representation for each sample and attempt to reconstruct a sample based on that embedding, often using Autoencoders~~\cite{Abati_2019_CVPR, Chong_2017, Hasan_2016_CVPR}. Predictive model based methods aim to model the current frame based on a set of previous frames, often relying on recurrent neural networks \cite{luo2017remembering, luo2017revisit, medel2016anomaly} or 3D convolutions~\cite{sabokrou2017deep, zhao2017spatio}. In some cases, reconstruction-based models are combined with prediction based methods for improved accuracy~\cite{zhao2017spatio}. In both cases, samples poorly reconstructed or predicted are considered anomalous. 

Generative models were also used to reconstruct, predict or model the distribution of the data, often using Variational Autoencoders (VAEs)~\cite{an2015variational} or GANs~\cite{akccay2019skip, lotter2015unsupervised, ravanbakhsh2017abnormal, ravanbakhsh2017training}.

A method proposed by Liu \etal~\cite{Liu_2018} uses a generative future frame prediction model 
and compares a prediction with its ground truth by evaluating differences in gradient-based features and optic flow. This method requires optic flow computation and 
generating a complete scene, which makes it costly and less robust to large scenery changes.

Recently, Morais~\etal~\cite{Morais_2019_CVPR} proposed an anomaly detection method using a fully connected RNN to analyze pose sequences. The method embeds a sequence, then uses reconstruction and prediction branches to generate past and future poses, respectively. Anomaly score is determined by the reconstruction and prediction errors of the model. 

\subsection{Graph Convolutional Networks}



To represent human poses as graphs, the inner-graph relations are described using weighted adjacency matrices. Each matrix could be static or learnable and represent any kind of relation.

In recent years, many approaches were proposed for applying deep learning based methods to graph data. Kipf and Welling~\cite{kipf2017semi} proposed the notion of {\it Fast Approximate Convolutions On Graphs}. Following Kipf and Welling, both temporal and multiple adjacency extensions were proposed.
Works by Yan~\etal~\cite{Yan2018SpatialTG} and Yu \etal~\cite{Yu_2018} proposed temporal extensions, with the former work proposing the use of separable spatial and temporal graph convolutions (ST-GCN), applied sequentially. We follow the basic ST-GCN block design, illustrated in Figure~\ref{fig:stgcn_diagram}.


{Veli{\v{c}}kovi{\'{c}}}~\etal~\cite{velickovic2018graph} proposed Graph Attention Networks, a GCN extension in which the weighting of neighboring nodes are inferred using an attention mechanism, relying on a fixed adjacency matrix only to determine neighboring nodes. 


Shi~\etal~\cite{2sagcn2019cvpr} recently extended the concept of spatio-temporal graph convolutions by using several adjacency matrices, of which some are learned or inferred. Inferred adjacency is determined using an embedded similarity measure, optimized during training. Adjacency matrices are summed prior to applying the convolution.

\subsection{Deep Clustering Models}

Deep clustering methods aim to provide useful cluster assignments by optimizing a deep model under a cluster inducing objective. For example, several recent methods jointly embed and cluster data using unsupervised representation learning methods, such as autoencoders, with clustering modules~\cite{caron2018deep, Dizaji_2017, Wang_2016, xie2016unsupervised}. 

A method proposed by Xie~\etal~\cite{xie2016unsupervised}, denoted {\it Deep Embedded Clustering (DEC),} proposed an alternating two-step approach. In the first step, a target distribution is calculated using the current cluster assignments. In the next step, the model is optimized to provide cluster assignments similar to the target distribution. Recent extensions tackled DEC's susceptibility to degenerate solutions using regularization methods and various post-processing means~\cite{Dizaji_2017, haeusser2018associative}.

\section{Method}

We design an anomaly detection algorithm that can operate in a number of different scenarios. The algorithm consists of a sequence of abstractions that are designed to help each step of the algorithm work better. First, we use a human pose detector on the input data.
This abstracts the problem and prevents the next steps from dealing with nuisance parameters such as viewpoint or illumination changes. 

Human actions are represented as space-time graphs and we embed (sub-sections~\ref{method:feature_extraction}, \ref{method:graph_conv}) and cluster (sub-section \ref{method:deep_clustering}) them in some latent space. Each action is now represented as a soft-assignment vector to a group of base actions. This abstracts the underlying type of actions (i.e., fine-grained or coarse-grained), leading to the final stage of learning their distribution. 
The tool we use for learning the distribution of soft-assignment vectors
is the Dirichlet process mixture (sub-section~\ref{method:normality_scoring}),
 and we fit a model to the data. 
This model is then used to determine if an action is normal or not.



\subsection{Feature Extraction}
\label{method:feature_extraction}

We wish to capture the relations between body joints, while at the same time provide robustness to external factors such as appearance, viewpoint and lighting. Therefore, we represent a person's pose with a graph.
\amir{new, please review}

Each node of the graph corresponds to a keypoint, a body joint, and each edge represents some relation between two nodes. Many keypoint relations exist, such as physical relations defined anatomically (e.g.\ the left wrist and elbow are connected) and action relations defined by movements that tend to be highly correlated in the context of a certain action  (e.g.\ the left and right knees tend to move in opposite directions while running). The directions of the graph rise from the fact that some relations are learned during the optimization process and are not symmetric. A nice bonus with this representation is being compact, which is very important for efficient video analysis.

In order to extend this formulation temporally, pose keypoints extracted from a video sequence are represented as a temporal sequence of pose graphs. The temporal pose graph is a time series of human joint locations. Temporal domain adjacency could be similarly defined by connecting joints in successive frames, allowing us to perform graph convolution operations exploiting both spatial and temporal dimensions of our sequence of pose graphs.

\begin{figure} [t!]  
\begin{tabular}{c}
\vspace{0.2cm}
\includegraphics[width=0.96\linewidth]{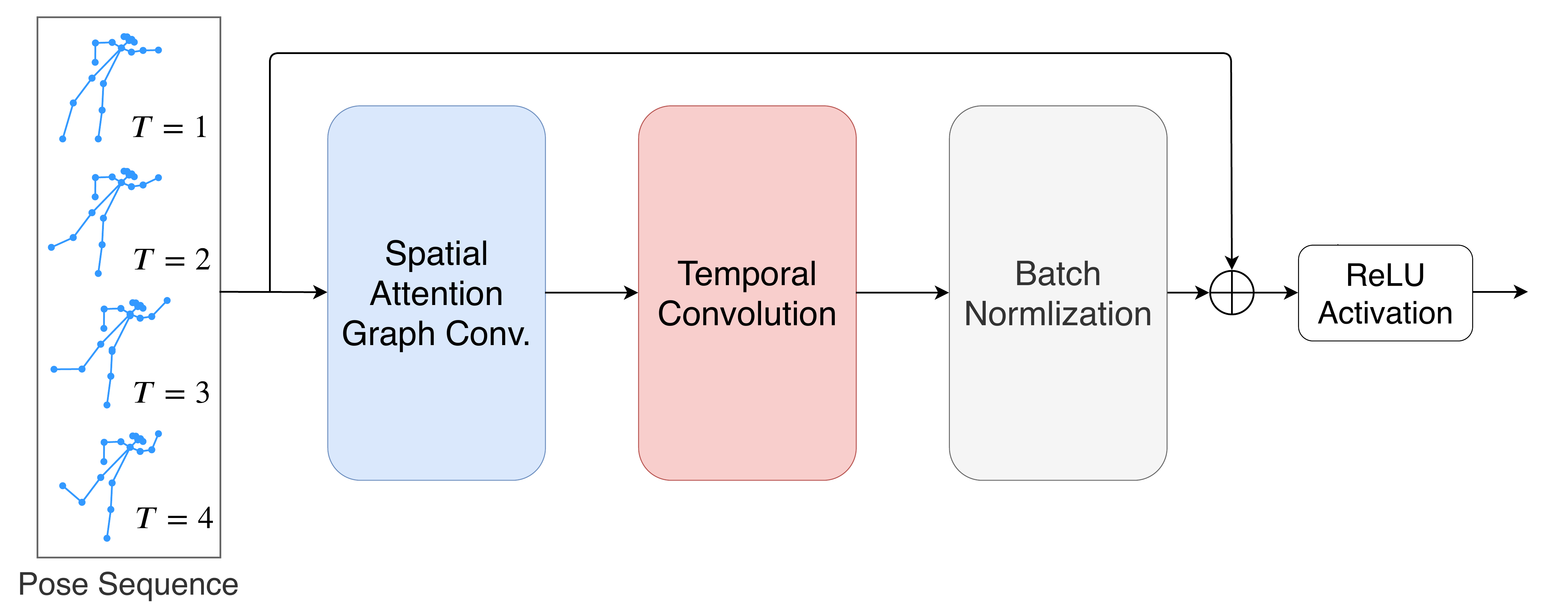} \vspace{0.2cm}
\vspace{-0.26cm}
\end{tabular}
\caption{{\bf Spatio-Temporal Graph Convolution Block:} The basic block used for constructing ST-GCAE. A spatial attention graph convolution (Figure \ref{fig:adpp_figure}) is followed by a temporal convolution and batch normalization. A residual connection is used.
}
\label{fig:stgcn_diagram}
\vspace{-0.42cm}  
\end{figure}

We propose a deep temporal graph autoencoder based architecture for embedding the temporal pose graphs. Building on the basic block design of ST-GCN, presented in Figure~\ref{fig:stgcn_diagram}, we substitute the basic GCN operator with a novel Spatial Attention Graph Convolution, presented next. 

We use this building block to construct a Spatio-Temporal Graph Convolutional Auto-Encoder, or \emph{ST-GCAE}. We use ST-GCAE to embed the spatio-temporal graph and take the  embedding to be the starting point for our clustering branch. 


\subsection{Spatial Attention Graph Convolution }
\label{method:graph_conv}
\begin{figure} [t!]  
\begin{tabular}{c}
\hspace{-0.0cm}
\includegraphics[width=0.94\linewidth]{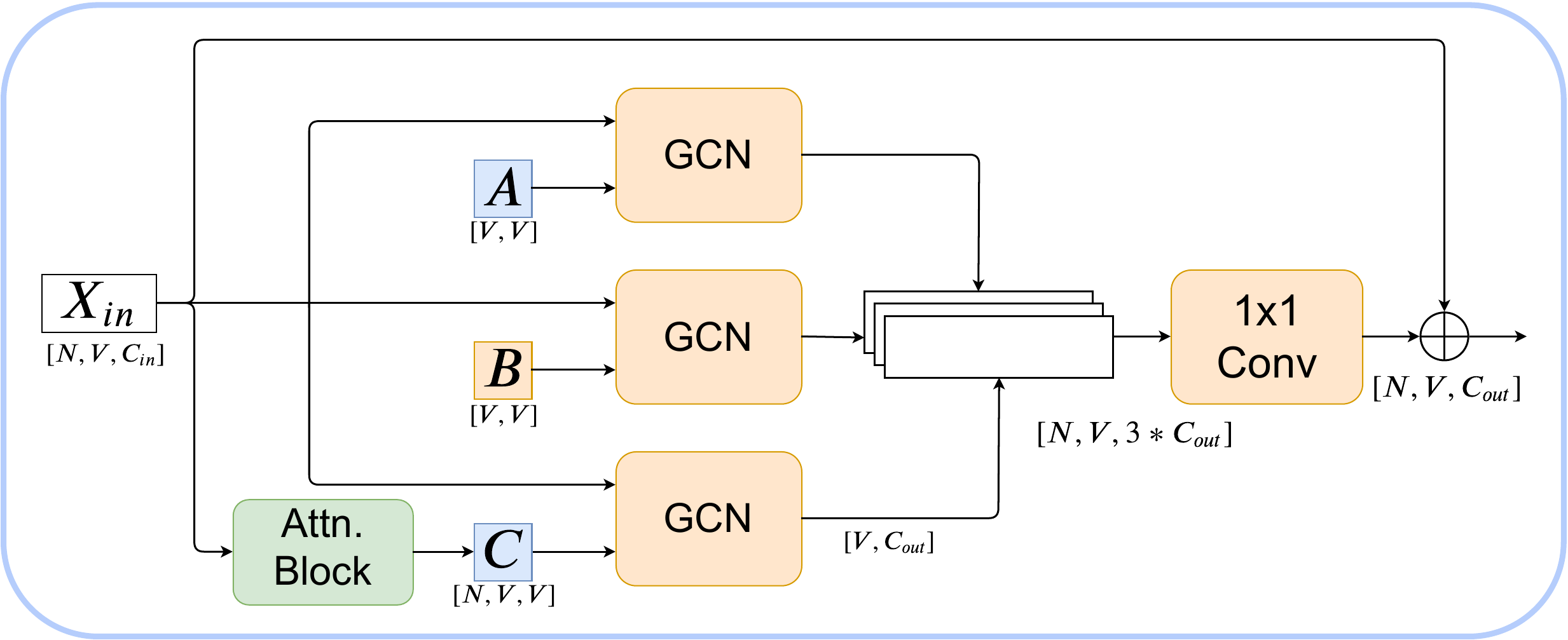} 
\vspace{-0.0cm}
\end{tabular}
\caption{{\bf Spatial-Attention Graph Convolution:} A zoom into 
our spatial graph convolving operator, comprised of three GCN~\cite{kipf2017semi} operators: One using a hard-coded physical adjacency matrix ($A$), the second using a global adjacency matrix learned during training ($B$), and the third using an adjacency matrix inferred using an attention submodule ($C$). A residual connection is used. GCN modules include batch normalization and ReLU activation, omitted for readability. }

\label{fig:adpp_figure}
\vspace{-0.4cm}
\end{figure}

We propose a new graph operator,  presented in Figure~\ref{fig:adpp_figure}, that uses adjacency matrices of three types: Static, Globally-learned and Inferred (attention-based). Each adjacency type is applied with its own GCN, using separate weights. The outputs from the GCNs are stacked in the channel dimension. A $1 \times 1$ convolution is applied as a learnable reduction measure for weighting the stacked outputs, and provides the required output channel number. 

The three adjacency matrices capture different aspects of the model: (i) The use of body-part connectivity as a prior over node relations, represented using the static adjacency matrix. (ii) Dataset level keypoint relations, captured by the global adjacency matrix, and (iii) Sample specific relations, captured by inferred adjacency matrices. Finally, the learnable reduction measure weights the different outputs. \amir{moved up and revised}


The static adjacency $A$ is fixed and shared by all layers. The globally-learnable matrix $B$ is learned individually at each layer, and applied equally to all samples during the forward pass.  The inferred adjacency matrices $C$ are based on an attention mechanism that uses learned weights to calculate a sample specific adjacency matrix, a different one for every sample in a batch. For example, for a batch of size~$N$ of graphs with~$V$ nodes, the inferred adjacency size is $[N, V, V]$, while other adjacencies are $[V, V]$ matrices.

 The globally-learned adjacency is learned by initializing a fully-connected graph, with a complete, uniform, adjacency matrix. The matrix is jointly optimized with the rest of the model's parameters during training. The computational overhead of this adjacency is small for graphs containing no more than a few dozen nodes.
 
 An inferred adjacency matrix is constructed using a graph self-attention layer. After evaluating a few attention models we chose a simple multiplicative attention mechanism. 
First, we embed the input twice, using two sets of learned weights. We then transpose one of the embedded matrices and take the dot product between the two and normalize. We then get the inferred adjacency matrix.  
The attention mechanism chosen is modular and may be replaced with other common alternatives.
 Further details are provided in the supplementary material. \amir{revised}

\subsection{Deep Embedded Clustering} \label{method:deep_clustering}
To build our dictionary of underlying actions, 
we take the training set samples and jointly embed and cluster them in some latent space. Each sample is then represented by its assignment probability to each of the underlying clusters. 
The objective is selected to provide distinct latent clusters, over which actions reside. \amir{revised}

We adapt the notion of \emph{Deep Embedded Clustering}~\cite{xie2016unsupervised} for clustering temporal graphs with our 
ST-GCAE architecture. The proposed clustering model consists of three parts, an encoder, a decoder, and a soft clustering layer.

Specifically, our ST-GCAE model maintains the graph's structure but uses large temporal strides with an increasing channel number to compress an input sequence to a latent vector. The decoder uses temporal up-sampling layers and additional graph convolutional blocks, for gradually restoring original channel count and temporal dimension.

The ST-GCAE's embedding is the starting point for clustering the data. 
The initial reconstruction based embedding is fine-tuned during our clustering optimization stage to reach the final clustering optimized embedding.

For each input sample~$\mathbf{x}_i$, we denote the encoder's latent embedding by~$\mathbf{z}_i$, and the soft cluster assignment calculated using the clustering layer by~$\mathbf{y}_i$. We denote the clustering layer's parameters by~$\Theta$. The probability~$p_{ik}$ for the $i$-th sample to be assigned to the $k$-th cluster is:

\vspace{-0.3cm}

\begin{equation}
\label{cluster_assignment}
        p_{ik} = Pr(y_i=k | \mathbf{z}_i, \Theta) = \frac{exp(\pmb{\theta}_k^T\mathbf{z}_i)}{\sum_{k'=1}^{K}exp(\pmb{\theta}_{k'}^T\mathbf{z}_i)}.
\end{equation}

We adopt the clustering objective and optimization algorithm proposed by~\cite{xie2016unsupervised}. The clustering objective is 
to minimize the KL divergence between the current model probabilistic clustering prediction ${P}$ and a target distribution ${Q}$:

\begin{equation}
\label{clustering_loss}
        L_{cluster} = KL({Q} || {P}) = \sum_{i}\sum_{k}q_{ik}log\frac{q_{ik}}{p_{ik}}.
        \vspace{0.01cm}
\end{equation}


The target distribution aims to strengthen current cluster assignments by normalizing and pushing each value closer to a 
value of either 0 or 1. Recurrent application of the function transforming ${P}$ to ${Q}$ will eventually result in a hard assignment vector. Each member of the target distribution is calculated using the following equation:

\begin{equation}
\label{target_dist}
        q_{ik} = \frac{p_{ik}/(\sum_{i'} p_{i'k})^{\frac{1}{2}}} {\sum_{k'} p_{ik'}/(\sum_{i'} p_{i'k'})^{\frac{1}{2}}}.
\end{equation}
\vspace{0.05cm}



The clustering layer is initialized by the K-means centroids calculated for the encoded training set. Optimization is done in 
Expectation-Maximization (EM) like fashion. During the Expectation step, the entire model is fixed and, the target distribution ${Q}$ is updated. During the Maximization stage, the model is optimized to minimize the clustering loss, $L_{cluster}$.

\subsection{Normality Scoring} \label{method:normality_scoring}

This model supports two types of multimodal distributions. One is at the cluster assignment level; the other is at the soft-assignment vector level. For example, an action may be assigned to more than one cluster (cluster-level assignment), leading to a multimodal soft-assignment vector. The soft-assignment vectors themselves (that capture actions) can be modeled by a multimodal distribution as well. \amir{reversed order between the first two paragraphs}

The Dirichlet process mixture model (DPMM) is a useful measure for evaluating the distribution of proportional data. It meets our required setup: (i) An estimation (fitting) phase,  during which a set of distribution parameters is evaluated, and (ii) An inference stage, providing a score for each embedded sample using the fitted model. A thorough overview of the model is given by Blei and Jordan~\cite{blei2006variational}.


The DPMM is a common mixture extension to the unimodal Dirichlet distribution and uses the Dirichlet Process, an infinite-dimensional extension of the Dirichlet distribution. This model is multimodal and able to capture each mode as a mixture component. A fitted model has several modes, each representing a set of proportions that correspond to one normal behavior. 
At test time, each sample is scored by its log probability using the fitted model. Further explanations and discussion on the use of DPMM are available in~\cite{blei2006variational, dinari2019distributed}.

\subsection{Training}

The training phase of the model consists of two stages, a pre-training stage for the autoencoder, in which the clustering branch of the network remains unchanged, and a fine-tuning stage in which both embedding and clustering are optimized. In detail: 

{\bf Pre-Training:} the model learns to encode and reconstruct a sequence by minimizing a reconstruction loss, denoted $L_{rec}$, which is an $L_2$ loss between the original temporal pose graphs and those reconstructed by ST-GCAE.

{\bf Fine-Tuning:} the model optimizes a combined loss function consisting of both the reconstruction loss and a clustering loss. Optimization is done such that the clustering layer is optimized w.r.t.\ $L_{cluster}$, the decoder is optimized w.r.t.\ $L_{rec}$ and the encoder is optimized w.r.t.\ both. The initialization of the clustering layer is done via $K$-means. As shown by \cite{Dizaji_2017}, while the encoder is optimized w.r.t.\ to both losses, the decoder is kept and acts as a regularizer for maintaining the embedding quality of the encoder. The combined loss for this stage is:
\begin{equation}
\label{combined loss}
        L_{combined} = L_{rec} + \lambda \cdot L_{cluster} .
\end{equation}

\section{Experiments}

\makeatletter
\renewcommand{\paragraph}{%
  \@startsection{paragraph}{4}%
  {\z@}{1.00ex \@plus 1ex \@minus .2ex}{-1em}%
  {\normalfont\normalsize\bfseries}%
}
\makeatother
We evaluated our model in two different settings, using three datasets. The first setting is the common video anomaly detection setting, which we denote as the \emph{Fine-grained} setting. In this setting, the normal sample consists of a single class and we seek to find fine-grained variations compared to it. For this setting, we use the \emph{ShanghaiTech Campus} dataset. The second is our new problem setting, which we denote \emph{Coarse-grained} anomaly detection, in which we seek to find abnormal actions that are different from those defined as normal. 


\subsection{ShanghaiTech Campus} \label{stc_dataset}

\paragraph{Dataset} The \emph{ShanghaiTech Campus dataset}~\cite{luo2017revisit} is one of the largest and most diverse datasets available for video anomaly detection. Presenting mostly person-based anomalies, it contains 130 abnormal events captured in 13 different scenes with complex lighting conditions and camera angles. Clips contain any number of people, from no people at all to over 20 people. The dataset contains over 300 untrimmed training and 100 untrimmed testing clips ranging from 15 seconds to over a minute long.

\paragraph{Experimental Setting} An experiment is comprised of two data splits, a training split containing \emph{normal} examples only and a test split containing both \emph{normal} and \emph{abnormal} examples. Training is conducted solely using the training split. A score is calculated for each frame individually, and the combined score is the area under ROC curve for the concatenation of all frame scores in the test set. 

We evaluate video streams of unknown length using a sliding-window approach. We split the input pose sequence to fixed-length, overlapping segments and score each individually. For clips with more than a single person, each person is scored individually. The maximal score over all the people in the frame is taken.
As the \textit{ShanghaiTech Campus} dataset is not annotated for pose, we use a 2D pose estimation model to extract human pose from every clip. 

We also evaluate our model using patch embeddings as input features instead of keypoint coordinates. Patches of pixel RGB data are cropped from around each keypoint. The patches are embedded using a CNN and patch feature vectors are used to embed each keypoint. All other aspects of the models are kept the same. 

Given the use of a pose estimation model, the patch embedding may be taken from one of the pose estimation model's hidden layers, requiring no additional computation compared to the coordinate-based variant, other than increased dimension for the input layer. Further details regarding this variant of our model, implementation, and the pose estimation method used are available in the supplemental material.

\paragraph{Evaluation} We follow the evaluation protocol of Luo~\etal~\cite{luo2017revisit} and report the \emph{Area under ROC Curve (AUC)} for our model in Table~\ref{tables:stc_results}. \emph{'Pose'} denotes the use of keypoint coordinates as the initial graph node embedding. \emph{'Patch'} denotes the use of patch embeddings vectors, as discussed in this section. Our model outperforms previous state of the art methods, both pose and pixel based, by a large margin.


\subsection{Coarse-Grained Anomaly Detection}


\subsubsection{Experimental Setting}
For our second setting of Coarse-Grained Anomaly Detection, a model is trained using a sample of a few action classes considered normal. Training is done without labels, in an unsupervised manner. The model is evaluated by its ability to tell whether a new unseen clip belongs to any of the actions that make up the normal sample. For this setting, we adopt two action recognition datasets to our needs. This gives us great flexibility and control over the type of normal/abnormal actions that we want to detect. The datasets are \emph{NTU-RGB+D} and \emph{Kinetics-250} that are provided with clip level action labels.

\begin{table}[!t]
\centering
\begin{tabular}{l c} 
\toprule

\multicolumn{2}{c}{\textbf{ShanghaiTech Campus}} \\
\midrule	

Luo \etal \cite{luo2017revisit} & 0.680  \\
Abati \etal \cite{Abati_2019_CVPR} & 0.725  \\
Liu \etal \cite{Liu_2018} & 0.728  \\ \midrule
Morais \etal \cite{Morais_2019_CVPR} & 0.734  \\ \midrule
Ours - Pose                      & \textbf{0.752} \\
Ours - Patches                   & \textbf{0.761} \\
\bottomrule
\end{tabular}
\vspace{0.1cm}		
\caption{ {\bf Fine-Grained Anomaly Detection Results:} Scores represent frame level AUC.
 \cite{Morais_2019_CVPR} uses keypoint coordinates as input.}
\vspace{-0.4cm}
\label{tables:stc_results}
\end{table}	
In this setting, we first select 3-5 action classes and denote them our \emph{split}. Classes are grouped into two sets of samples, \emph{split} samples, and \emph{non-split} samples. All labels are dropped. No labels are used beyond this point, except for the final evaluation phase.

We conduct two complementary experiments. {\it Few vs. Many} where there are few normal actions (say 3-5) in the training set and many (tens or even hundreds) actions that are denoted abnormal in the test set. We then repeat the experiment but switch roles of the train and test sets and denote this as {\it Many vs. Few}. 

We repeat the above experiments for two types of splits. The first kind, termed {\it random splits}, is made of sets of 3-5 classes selected at random from each dataset. The second, which we call {\it meaningful splits}, is made of action splits that are subjectively grouped following some binding logic regarding the action's physical or environmental properties. A sample of meaningful and random splits is provided in Table \ref{table:meaningful_splits}. We use 10 random and 10 meaningful splits for evaluating each dataset.

\begin{table*}[!t]
\centering
\begin{tabular}{l cccc cccc} 
\toprule

 &  \multicolumn{4}{c}{\textbf{NTU-RGB+D}} & \multicolumn{4}{c}{\textbf{Kinetics-250}} \\
\cmidrule(lr){2-5} \cmidrule(lr){6-9}
\multicolumn{1}{l}{} & \multicolumn{2}{c}{{Few vs.\ Many}} & \multicolumn{2}{c}{{Many vs.\ Few}} & \multicolumn{2}{c}{{Few vs.\ Many}} & \multicolumn{2}{c}{{Many vs.\ Few}} \\
\cmidrule(lr){2-3} \cmidrule(lr){4-5} \cmidrule(lr){6-7} \cmidrule(lr){8-9}
\textbf{Method} & \textit{Random} & \textit{Meaningful} & \textit{Random} & \textit{Meaningful} & \textit{Random} & \textit{Meaningful} & \textit{Random} & \textit{Meaningful} \\
\midrule	

Supervised         & \underline{0.86} & \underline{0.83} & 
 \underline{0.82} & \underline{0.90} &
 \underline{0.77} & 0.71 &
 \underline{0.63} & \underline{0.82}  \\
\midrule            
Rec. Loss                
& 0.50 & 0.54  
& 0.53 & 0.54  
& 0.45 & 0.46  
& 0.51 & 0.61  \\
OC-SVM                    
& 0.60 & 0.67 
& 0.60 & 0.69 
& 0.56 & 0.56 
& 0.52 & 0.47 \\ 
Liu \etal \cite{Liu_2018} 
& 0.57 & 0.64 
& 0.56 & 0.63 
& 0.55 & 0.60 
& 0.55 & 0.58 \\
Morais \etal \cite{Morais_2019_CVPR} & - & - & - & - 
& 0.57 & 0.59 
& 0.56 & 0.58 \\
Ours                      
& \textbf{0.73} & \textbf{0.82} 
& \textbf{0.72} & \textbf{0.85} 
& {\textbf{0.65}} & {\underline{\textbf{0.73}}}  
& {\textbf{0.62}} & {\textbf{0.74}} \\
\bottomrule
\end{tabular}
\vspace{0.08cm}		
\caption{ {\bf Coarse-Grained Experiment Results:} Values represent area under the ROC curve (AUC). In bold are the results of the best performing \emph{unsupervised} method. Underlined is the best method of all. For all experiments $K=20$ clusters, see section~\ref{method:deep_clustering} for details. 
It~should be noted that AUC=0.50 in case of random choice.
}
\vspace{-0.3cm}
\label{tables:results}
\end{table*}	
\vspace{-0.2cm}

\subsubsection{Methods Evaluated}
We compare our algorithm to several anomaly detection algorithms. All algorithms but the last one are unsupervised:

\paragraph{Autoencoder reconstruction loss}
We use the reconstruction loss of our ST-GCAE model.  
In all experiments, the ST-GCAE reached convergence prior to the deep clustering fine-tuning stage. Further optimization of the ST-GCAE yielded no consistent improvement in results.
\paragraph{Autoencoder based one-class SVM}
We fit a one-class SVM model using the encoded pose sequence representations (denoted $\mathbf{z}_i$ in section \ref{method:deep_clustering}). During test time, the corresponding representation of each sample is scored using the fitted model.
\paragraph{Video anomaly detection methods} 
We train the \emph{Future Frame Prediction} model proposed by Liu~\etal~\cite{Liu_2018} and the \emph{Skeleton Trajectory} model proposed by Morais~\etal~\cite{Morais_2019_CVPR} using our various dataset splits. Anomaly scores for each video are obtained by averaging the per-frame scores provided by the model. As the method proposed by Morais~\etal only handles 2D pose, it was not applied to the 3D annotated NTU dataset. 

\paragraph{Classifier softmax scores}
The \textit{supervised} baseline uses a classifier trained to classify each of the classes from the dataset split. The classifier architecture is based on the one proposed by~\cite{Yan2018SpatialTG}. To handle the significantly smaller number of samples, we use a shallower variant. For classifier architecture and implementation details, see suppl. 

During the evaluation phase, a sample is passed through the classifier and its softmax output values are recorded. Anomaly score in this method is calculated by either using the softmax vector's max value or by using the Dirichlet normality score from section \ref{method:normality_scoring}, using softmax probabilities as input. We found Dirichlet based scoring to perform better for most cases, and we report results based on it.

It is important to note that this method is fundamentally different from our method and the other baselines. The classifier based method is a {\it supervised} method, relying on class action labels that were not used by other methods. It is thus not directly comparable and is here for reference only.

\begin{table}[h] 
\centering
{\small \begin{tabular}{c l} 
\toprule

\multicolumn{2}{c}{\textbf{Kinetics}} \\
\midrule            
Random 1 & Arm wrestling (6), Crawling baby (77) \\ & Presenting weather forecast (254), \\ & Surfing crowd (336) \\ \midrule
Dancing & Belly dancing (18), Capoeira (43),\\ & Line dancing (75), Salsa (283), \\ &Tango (348), Zumba (399) \\ \midrule
Gym & Lunge (183), Pull Ups (255), Push Up (260),\\ & Situp (305), Squat (330) \\ 
\bottomrule \toprule
\multicolumn{2}{c}{\textbf{NTU-RGB+D}} \\ \midrule

Office & Answer phone (28), Play with phone/tablet (29), \\ & Typing on a keyboard (30), Read watch (33) \\ 
\midrule
Fighting  & Punching (50), Kicking (51), Pushing (52), \\ & Patting on back (53) \\

\bottomrule
\end{tabular}}
\vspace{0.08cm}		
\caption{\small{\textbf{Split Examples:} A subset of the random and meaningful splits used for evaluating {\it Kinetics} and {\it NTU-RGB+D} datasets. For each split we list the included classes. Numbers in parentheses are the numeric class labels. 
For a complete list, See suppl. 
}}
\vspace{-0.4cm}
\label{table:meaningful_splits}
\end{table}	  
\subsubsection{Datasets}

\paragraph{NTU-RGB+D} The {\it NTU-RGB+D} dataset by Shahroudy \etal~\cite{Shahroudy_2016_CVPR} consists of clips showing one or two people performing one of 60 action classes. Classes include both actions of a single person and two-person interactions, captured using static cameras. It is provided with 3D joint measurements that are estimated using a Kinect depth sensor.

For this dataset, we use a model configuration similar to the one used for the \emph{ShanghaiTech} experiments, with dimensions adapted for 3D pose.

\paragraph{Kinetics-250}

The {\it Kinetics} dataset by Kay \etal~\cite{kay2017kinetics} is a collection of 400 action classes, each with over 400 clips that are 10 seconds long. The clips were downloaded from YouTube and may contain any number of people that are not guaranteed to be fully visible. 

Since Kinetics was not intended originally for pose estimation, some classes are unidentifiable by human pose extraction methods, e.g., the {\it hair braiding} class contains mostly clips focused on arms and heads. For such videos, a full-body pose estimation algorithm will yield zero keypoints for most cases. 

Therefore, we use a subset of {\it Kinetics-400} that is suitable for evaluation using pose sequences. To do that, we turn to the action classification results of~\cite{Yan2018SpatialTG}. Using their publicly available model we pick a subset of the 250 best-performing action classes, ranked by their {\it top-1} training classification accuracy. The accuracy of the class that had the lowest score is~$18\%$. We denote our subset {\it Kinetics-250}.



Due to the vast size of Kinetics ($\sim$1000x larger than ShanghaiTech), 
we used a single GCN for the spatial convolution,
using static $A$ adjacency matrices only, and no pooling. This makes this block identical to the one proposed by~\cite{Yan2018SpatialTG}, used for this specific setting \emph{only}. We quantify the degradation of this variant in the suppl. 
\textit{Kinetics} is not annotated for pose and we use a 2D pose estimation model.


\subsubsection{Evaluation} 

We report \emph{Area under ROC Curve (AUC)} results in Table~\ref{tables:results}. As these datasets require clip level annotations, the sliding window approach is not required for our method, and each temporal pose graph is evaluated in a single forward pass, with the highest scoring person taken.

As can be seen, our algorithm outperforms all four competing (unsupervised) methods, often by a large margin. The algorithm works well in both random and meaningful split modes, as well as in the Few vs.\ many and Many vs.\ few settings. Observe, however, that the algorithm works better on the meaningful splits (compared to the random splits). We believe this is because meaningful splits share similar patterns. 

The table also reveals the impact of the quality of pose estimation on results. That is, the {\it NTU-RGB+D} dataset is cleaner and the human pose is recovered using the Kinect depth sensor. As a result, the estimated poses are more accurate and the results are generally better than the {\it Kinetics-250} dataset.



\vspace{0.2cm}
\subsection{Fail Cases} \label{results:fails_cases}
\begin{figure}
\begin{center}
\begin{tabular}{@{\hskip 0in}c@{\hskip 0.05in}c@{\hskip 0.05in}c@{\hskip 0.06in}@{\hskip 0in}}
\includegraphics[width=0.31\linewidth]{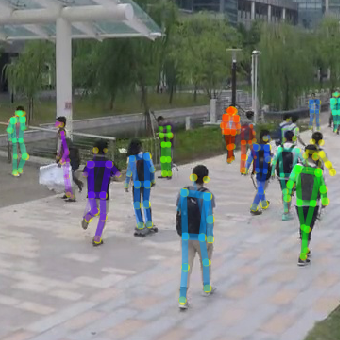}  & 
\includegraphics[width=0.31\linewidth]{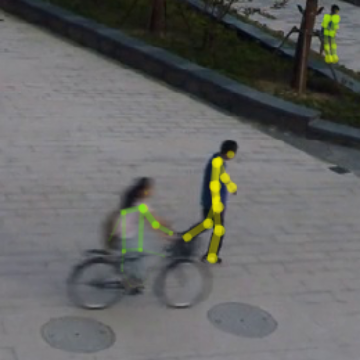} & 
\includegraphics[width=0.31\linewidth]{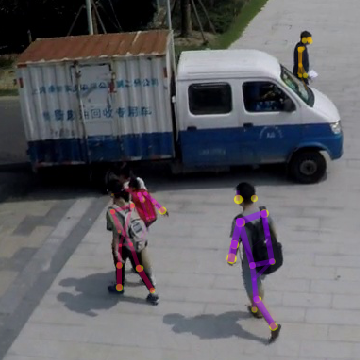} \\ 

\multicolumn{1}{c}{(a)} & \multicolumn{1}{c}{(b)} & \multicolumn{1}{c}{(c)} \vspace{-0.25cm}\\
\end{tabular}
\end{center}
\caption{{\small \textbf{Failure Cases, \textit{ShanghaiTech}:} Frames overlayed with extracted pose. In Column (a), the large crowd is occluding the abnormal skater and each other causing multiple misses. Column (b) depicts a cyclist, considered abnormal. Fast movement caused pose estimation failure, preventing detection. Column (c) depicts a vehicle in the frame, which is not addressed by our method.  
}
\vspace{-0.1cm}
\label{fig:fail_cases_stc}}
\end{figure}

Figure~\ref{fig:fail_cases_stc} shows some failure cases. The recovered pose graph is superimposed on the image. As can be seen, there is significant variability in scenes, viewpoints and poses of the people in a single clip. Depicted in column (a), a highly crowded scene causes numerous occlusions and people being partially detected. The large number of partially extracted people causes a large variation in model provided scores, and misses the abnormal skater for multiple frames.

The two failures depicted in columns (b-c) show the weakness of relying on extracted pose for representing actions in a clip. Column (b) shows a cyclist is very partially extracted by the pose estimation method and missed by the model. Column (c) shows a non-person related event, not handled by our model. Here, a vehicle crossing the frame.




\subsection{Ablation Study} \label{results:ablation}

We conduct a number of experiments to evaluate the robustness of our model to noisy normal training sets, i.e.,\ having some percentage of abnormal actions present in the training set, presented next. We also conduct experiments to evaluate the importance of key model components and the stages of our clustering approach, presented in the suppl.

\begin{figure} [t!]  
\begin{tabular}{c}
\hspace{-0.3cm}
\includegraphics[width=1.0\linewidth]{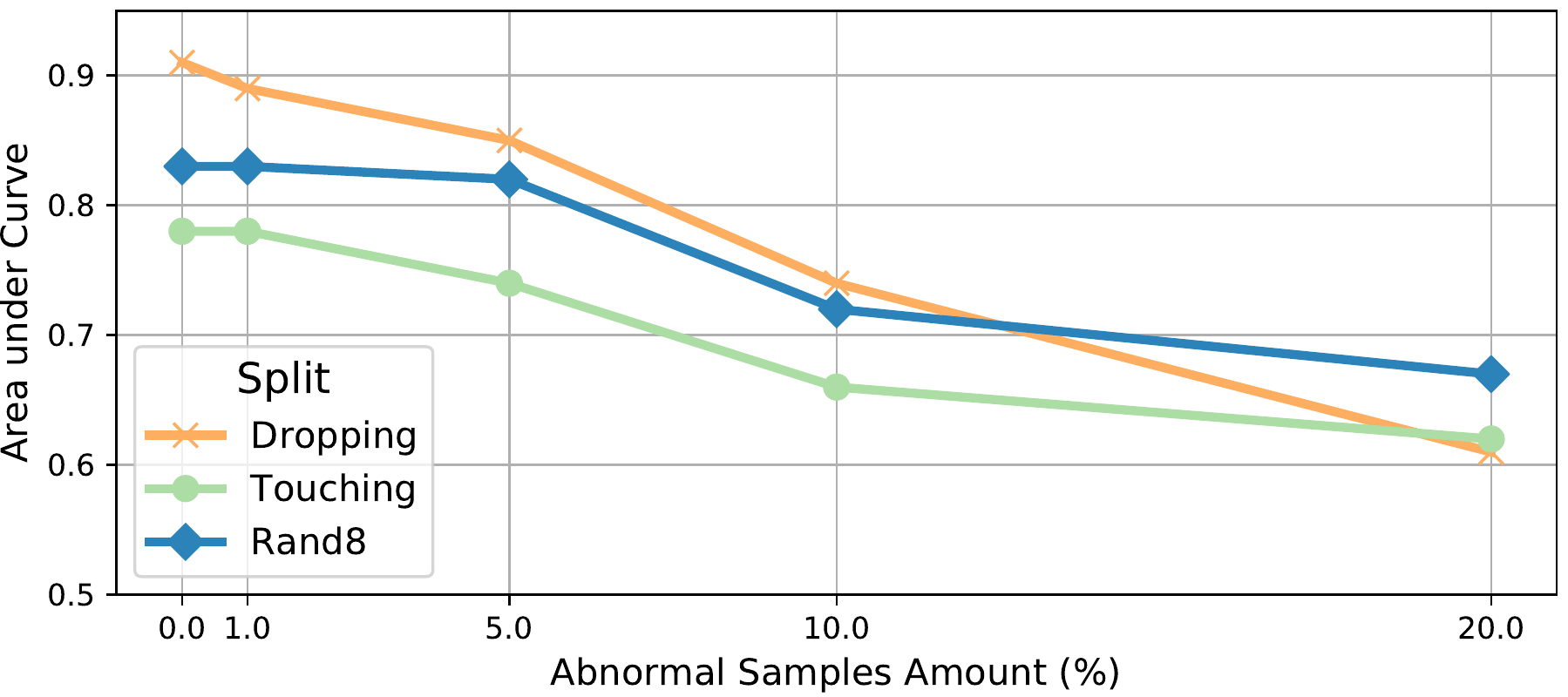} \vspace{0.1cm}
\end{tabular}
\caption{{\bf AUC Loss for Training with Noisy Data:} Performance of models trained for NTU-RGB+D splits when a percentage of abnormal samples are added at random. The model is robust to significant amount of noise. At $20\%$, noise surpasses the amount of data for some of the underlying classes making up the split. 
Different curves denote different dataset splits.
}

\label{fig:noisy_data}
\vspace{-0.3cm}
\end{figure}

\paragraph{Robustness to Noise} In many scenarios, it is impossible to determine whether a dataset contains nothing but normal samples, and some robustness to noise is required. To evaluate the model's robustness to the presence of abnormal examples in the normal training sample, we introduce a varying number of abnormal samples chosen at random to the training set. These are taken from the unused abnormal portion of the dataset. Results are presented in Figure~\ref{fig:noisy_data}. Our model is robust and handles large amount of abnormal data during training with little performance loss. 

For most anomaly detection settings, events occurring at a $5\%$ rate are considered very frequent. Our model loses on average less than $10\%$ of performance when trained with this amount of distractions. When trained with $20\%$ abnormal noise, there is a considerable decline in performance. In this setting, the training set usually consists of 5 classes, so $20\%$ distraction rate may be larger than an individual underlying class.

\vspace{0.3cm}


\section{Conclusion}
We propose an anomaly detection algorithm that relies on estimated human poses. The human poses are represented as temporal pose graphs and we jointly embed and cluster them in a latent space.
As a result, each action is represented as a soft-assignment vector in latent space. We analyze the distribution of these vectors using the Dirichlet Process Mixture Model. The normality score provided by the model is used to determine if the action is normal or not.

The proposed algorithm works on both fine-grained anomaly detection, where the goal is to detect variations of a single action (e.g.,\ walking), as well as a new coarse-grained anomaly detection setting, where the goal is to distinguish between normal and abnormal actions.

Extensive experiments show that we achieve state-of-the-art results on ShanghaiTech, one of the leading (fine-grained) anomaly detection data sets. We also outperform existing unsupervised methods on our new coarse-grained anomaly detection test.

{\small
\bibliographystyle{ieee_fullname}
\bibliography{egbib}
}

\clearpage

\appendix

\section{Supplementary Material}

The supplementary material provides additional ablation experiments, as well as details regarding experiment splits and results, and information regarding the proposed spatial attention graph convolution, and implementation of our method and of the baseline methods.

Specifically, Section~\ref{supp_ablation} presents further ablation experiments conducted to evaluate our model. Section~\ref{supp_basewords} 
presents the base action-words learned by our model in both settings. 

In Section~\ref{supp_sagc} we go into further details regarding the proposed spatial attention graph convolution operator. Section~\ref{supp_impl} provides implementation details for our method, and Section~\ref{supp_baseline_impl} describes the implementations the of baseline methods used. 

For the \emph{Coarse-grained} experiments, per-split results and class lists are available in Section~\ref{supp_results} and in Section~\ref{supp_split_list} respectively. Finally, the complete list of classes used for \textit{Kinetics-250} is provided in Section~\ref{supp_kin250}. 



    
    
    
\section{Ablation Experiments - Cont.} \label{supp_ablation}

In this section we provide further ablation experiments used to evaluate different model components:

\paragraph{Input and Spatial Convolution} 
In Table~\ref{tables:part_ablation} we evaluate the contribution of two key components of our configuration. First, the input representation for nodes. We compare the \emph{Pose} and \emph{Patch} keypoint representations.

In the \emph{Pose} representation, each graph node is represented by its coordinate values ($[x, y, conf.]$) provided by the pose estimation model. In the \emph{Patch} representation, we use features extracted using a CNN from a patch surrounding each keypoint. 

Then, we evaluate the spatial graph operator used. We deonote our spatial attention graph convolution by \emph{SAGC}, and the single adjacency variant by \emph{GCN}.
It is evident that both the use of patches and of the spatial attention graph convolution play a key role in our results. 

\label{supp:clustering_ablation}
\paragraph{Clustering Components} We conducted a number of ablation tests on one of the splits to measure the importance of the number of clusters $K$, the clustering initialization method, the proposed normality score, and the fine-tuning training stage. Results are summarized in Table \ref{tables:clustering_ablation}.

The different columns correspond to different numbers of clusters. As can be seen, best results are usually achieved for $K=20$ and we use that value through all our experiments in the coarse setting.
Each pair of rows correspond to two normality scores that we evaluate. "Dir." stands for the Dirichlet based normality score. "Max" simply takes the maximum value of the softmax layer, the soft-assignment vector. Our proposed normality score performs consistently better (except for the case of $K=5$).

The first two rows of the table evaluate the importance of initializing the clustering layer. Rows 3-4 show the improvement gained by using $K$-means for initialization compared to the random initialization used in rows 1-2. 

Next, we evaluate the importance of the fine-tuning stage. Models that were fine-tuned are denoted by \textit{DEC} in the table. Models in which the fine-tuning stage was skipped are denoted by \textit{No DEC}. Rows 3-4 show results without using the fine-tuning stage, while rows 5-6 show results with. As can be seen, results improve considerably (except for the case of $K=5$).


\begin{table}
\centering
\begin{tabular}{l c c} 
\toprule
\textbf{Method} & {GCN} & {SAGC} \\ 
\midrule            
Pose Coords.      & 0.750 & 0.753 \\ 
Patches           & 0.749 & \textbf{0.761} \\ 
\bottomrule
\end{tabular}
\vspace{0.1cm}		
\caption{ \small{\bf Input and Spatial Convolution:} 
Results of different model variations for the {\it ShanghaiTech Campus} dataset. Rows denote input node representations, \emph{Pose} for keypoint coordinates, \emph{Pathces} for surrounding patch features. Columns denotes different spatial convolutions: \emph{GCN} uses the physical adjacency matrix only. \emph{SAGC} is our proposed operator.
\emph{SAGC} provides a meaningful improvement when used with patch embedding. Values represent frame level AUC.
}
\label{tables:part_ablation}
\end{table}	

\begin{table}
\centering
\begin{tabular}{l c c c} 
\toprule
\textbf{Method} & \textbf{5} & \textbf{20} & \textbf{50} \\
\midrule            
Random init, DEC, Max           & 0.45 & 0.42 & 0.44 \\ 
Random init, DEC, Dir.               & 0.48 & 0.52 & 0.49 \\ \midrule
K-means init, No DEC, Max  & 0.57 & 0.51 & 0.48 \\ 
K-means init, No DEC, Dir.      & 0.51 & 0.59 & 0.57 \\ \midrule
K-means init, DEC, Max	   & 0.58 & 0.71 & 0.72 \\ 
K-means init, DEC, Dir. 	       & 0.68 & \bf{0.82} & 0.74 \\
\bottomrule
\end{tabular}
\vspace{0.1cm}		
\caption{ \small{\bf Clustering Components:} Results for {\it Kinetics-250} Few vs.\ Many Experiment, split "Music": Values represent Area under RoC curves. Column values represent the number of clusters. ``Max'' / ``Dir.'' denotes the normality scoring method used, the maximal softmax value or Dirichlet based model. Values represent frame level AUC.
See section~\ref{supp:clustering_ablation} for further details.}
\label{tables:clustering_ablation}
\vspace{-0.3cm}
\end{table}	
\section{Visualization of Action-words} \label{supp_basewords}

\begin{figure}
\begin{center}
\begin{tabular}{@{\hskip 0in}c@{\hskip 0.05in}c@{\hskip 0.05in}c@{\hskip 0.06in}@{\hskip 0in}}

\includegraphics[width=0.3\linewidth]{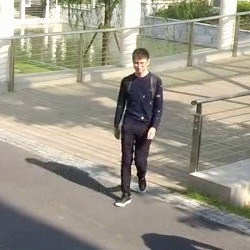} & 
\includegraphics[width=0.3\linewidth]{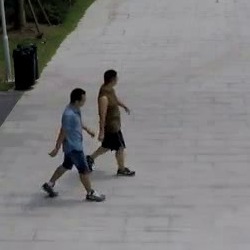} & 
\includegraphics[width=0.3\linewidth]{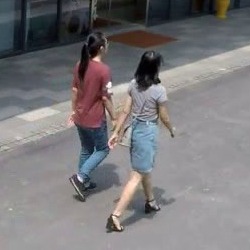} \vspace{-0.1cm} \\
\multicolumn{1}{c}{(a)} & \multicolumn{1}{c}{(b)} & \multicolumn{1}{c}{(c)} \vspace{0.0cm}\\

\includegraphics[width=0.3\linewidth]{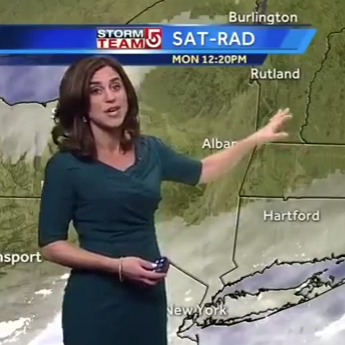} & 
\includegraphics[width=0.3\linewidth]{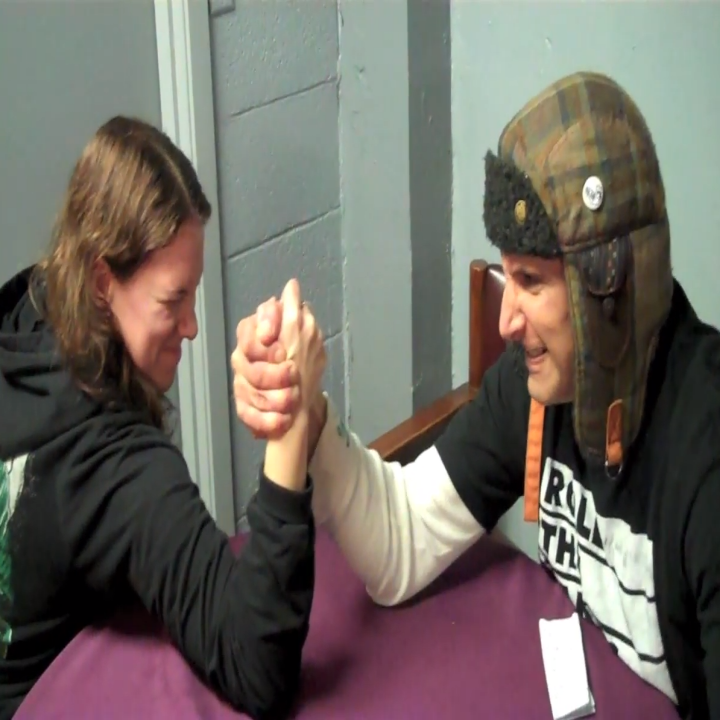} & 
\includegraphics[width=0.3\linewidth]{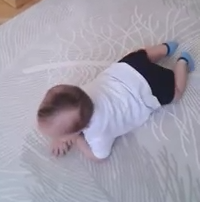} \\
\multicolumn{1}{c}{(d)} & \multicolumn{1}{c}{(e)} & \multicolumn{1}{c}{(f)} \vspace{-0.25cm}\\

\end{tabular}
\end{center}

\caption{{\small \textbf{Base Words:} 
Samples closest to different cluster centroids, extracted from a trained model. 
Top: Fine-grained. \emph{ShanghaiTech} model action words.
Bottom: Coarse-grained. Actions words from a model trained on \emph{Kinetics-250}, split \emph{Random 6}.
For the \emph{fine-grained} setting, clusters capture small variations of the same action. However, for \emph{coarse-grained}, actions close to the centroids vary considerably, and depict an essential manifestation of underlying actions depicted. 
}}

\label{supp_basewords_combined}
\end{figure}

It is instructive to look at the clusters of the different data sets (Figure~\ref{supp_basewords_combined}). Top row shows some cluster centers in the \emph{fine-grained} setting and bottom row shows some cluster centers in the \emph{coarse-grained} setting. As can be seen, the variation in the fine grained setting is mainly due to viewpoint, because most of the actions are variation on walking.
On the other hand, the variability of the coarse-grained data set demonstrate the large variation in the actions that handled by our algorithm.


\paragraph{Fine-grained} In this setting,  
actions close to different cluster centroids depict common variations of the singular action taken to be normal, in this case, walking directions. The dictionary action words depict clear, unoccluded and full body samples from normal actions.

\paragraph{Coarse-grained} Frames selected from clips corresponding to base words extracted from a model trained on the \emph{Kinetics-250} dataset, split \emph{Random 6}. Here, actions close to the centroids depict an essential manifestation of underlying action classes depicted. Several clusters in this case depict the underlying actions used to construct the split:
Image~(d) shows a sample from the 'presenting weather' class. Facing the camera, pointing at a screen with the left arm while keeping the right one mostly static is highly representative of presenting weather; 
Image~(e) depicts the common pose from the 'arm wrestling' class  and,
Image~(f) does the same for the 'crawling' class.

\section{Spatial Attention Graph Convolution} \label{supp_sagc}

We will now present in detail several components of our spatial attention graph convolution layer. It is important to note that every kind of adjacency is applied independently, with different convolutional weights. After concatenating outputs from all GCNs, dimensions are reduced using a learnable $1\times1$ convolution operator. 


For this section, $N$ denotes the number of samples, $V$ is the number of graph nodes and $C$ is the number of channels. 
During the spatial processing phase, pose from each frame is processed independently of temporal relations. 

\paragraph{GCN Modules} 
We use three GCN operators, each corresponding to a different kind of adjacency matrices. Following each GCN we apply batch normalization and a ReLU activation. If a single adjacency matrix is provided, as in the static and globally-learnable cases, it is applied equally to all inputs. In the inferred case, 
every sample is applied the corresponding adjacency matrix.


\paragraph{Attention Mechanism} Generally, 
the attention mechanism is modular and can be replaced by any graph attention model meeting the same input and output dimensions.
There are several alternatives (\cite{vaswani2017attention, velickovic2018graph}) which come at significant computational cost. We chose a simpler mechanism, inspired by~\cite{Luong_2015, 2sagcn2019cvpr}. Each sample's node feature matrix, shaped $[V, C_{in}]$, is multiplied by two separate attention weight matrices shaped $[C_{in}, C_{attn}]$. One is transposed and the dot product between the two is taken, followed by normalization. We found this simple mechanism to be useful and powerful.
\section{Implementation Details} \label{supp_impl}

\paragraph{Pose Estimation} For extracting pose graphs from the \emph{ShanghaiTech} dataset we used Alphapose~\cite{fang2017rmpe}. Pose tracking is done using Poseflow~\cite{xiu2018poseflow}. Each keypoint is provided with a confidence value. For \emph{Kinetics-250} we use the publicly available keypoints\footnote{\url{https://github.com/open-mmlab/mmskeleton}} extracted using Openpose\cite{Cao_2017}.
The above datasets use 2D keypoints with confidence values.

The \emph{NTU-RGB+D} dataset is provided with 3D keypoint annotations, acquired using a Kinect sensor. For 3D annotations, there are 25 keypoints for each person. 

\paragraph{Patch Inputs} The \emph{ShanghaiTech} model variant using patch features as input network embeddings works as following: First, a pose graph is extracted. Then, around each keypoint in the corresponding frame, a $16\times16$ patch is cropped. Given that pose estimation models rely on object detectors (Alphapose uses FasterRCNN\cite{ren2015faster}), intermediate features from the detector may be used with no added computation. For simplicity, we embedded each patch using a publically available \emph{ResNet} model\footnote{\url{https://github.com/akamaster/pytorch_resnet_cifar10}}. Features used as input are the $64$ dimensional output of the global average pooling layer. Other than the input layer's shape, no changes were made to the network.

\paragraph{Architecture} A symmetric structure was used for ST-GCAE. 
Temporal downsampling by factors of 2 and 3 were applied in the second and forth blocks. The decoder is symmetrically reversed. 
We use~${K=20}$ clusters for \emph{NTU-RGB+D} and \emph{Kinetics-250} and~${K=10}$ clusters for \emph{ShanghaiTech}. During the training stage, the samples were augmented using random rotations and flips. During the evaluation we average results for each sample over its augmented variants. Pre- and post-processing practices were applied equally to our method and all baseline methods. 

\paragraph{Training} Each model begins with a pre-training stage, where the clustering loss isn't used. A fine-tuning stage of roughly equal length follows during which the model is optimized using the combined loss, with the clustering loss coefficient~$\lambda = 0.5$ for all experiments. The {\it Adam} optimizer~\cite{kingma_adam} is used.

\section{Baseline Implementation Details} \label{supp_baseline_impl}
\makeatletter
\renewcommand{\paragraph}{%
  \@startsection{paragraph}{4}%
  {\z@}{1.00ex \@plus 1ex \@minus .2ex}{-1em}%
  {\normalfont\normalsize\bfseries}%
}
\makeatother
\paragraph{Video anomaly detection methods}
The evaluation of the future frame prediction method by Liu~\etal~\cite{Liu_2018} was conducted using their publicly available implementation\footnote{\url{https://github.com/stevenliuwen/ano\_pred\_cvpr2018/}}.
Similarly, the evaluation of the Trajectory based anomaly detection method by Morais~\etal~\cite{Morais_2019_CVPR} was also conducted using their publicly available implementation\footnote{\url{https://github.com/RomeroBarata/skeleton_based_anomaly_detection}}.
Training was done using default parameters used by the authors, and changes were only made to adapt the data loading portion of the models to our datasets.

\paragraph{Classifier softmax scores}
The classifier based supervised baseline used for comparison is based on the basic ST-GCN block used for our method. We use a model based on the architecture proposed by Yan~\etal~\cite{Yan2018SpatialTG}, using their implementation\footnote{\url{https://github.com/yysijie/st-gcn/}}. For the \textit{Few vs.\ Many} experiments we use 6 ST-GCN blocks, two with 64 channel outputs, two with 128 channels and two with~256. This is the smaller model of the two, designed for the smaller amount of data available for the \textit{Few vs.\ Many} experiments. For the \textit{Many vs.\ Few} experiments we use 9 ST-GCN blocks, three with~64 channel outputs, three with~128 channels and three with~256. Both architectures use residual connections in each block and a temporal kernel size of 9.  In both, the last layers with 64 and 128 channels perform temporal downsampling by a factor of 2. 
Training was done using the \textit{Adam} optimizer.

The method provides a probability vector of per-class assignments. The vector is used as the input to the Dirichlet based normality scoring method that was used by our model. The scoring function's parameters are fitted using the training set data considered ``normal'', and in test time, each sample is scored using the fitted parameters.


\section{Detailed Experiment Results} \label{supp_results}

Detailed results are provided for each dataset, method and setting. Results for \textit{NTU-RGB+D} are provided in page~\pageref{tables:supp-results-ntu}
and for \textit{Kinetics-250} in page~\pageref{tables:supp-results-kinetics}. 


We use~\emph{``sup.''} to denote the supervised, classifier-based baseline in all figures. This method is fundamentally different from all others, and uses the class labels for supervision.

One can observe that for all settings our method is the top performer in most splits compared to unsupervised methods, often by a large margin. 

\section{Class Splits Table} \label{supp_split_list}

The list of random and meaningful splits used for evaluation is available in Table \ref{table:supp_splits_ntu} for \textit{NTU-RGB+D} and Table \ref{table:supp_splits_kinetics} for \textit{Kinetics-250}.

Random splits were used to objectively evaluate the ability of a model to capture a specific subset of unrelated actions. Meaningful splits were chosen subjectively to contain a binding logic regarding the action's physical or environmental properties, e.g.\ actions depicting musicians playing or actions one would likely see in a gym.

\begin{figure*} [!h]	
\hspace*{-0.0cm}\begin{tabular}{c}
\hspace{1.12cm}\includegraphics[width=0.84\textwidth]{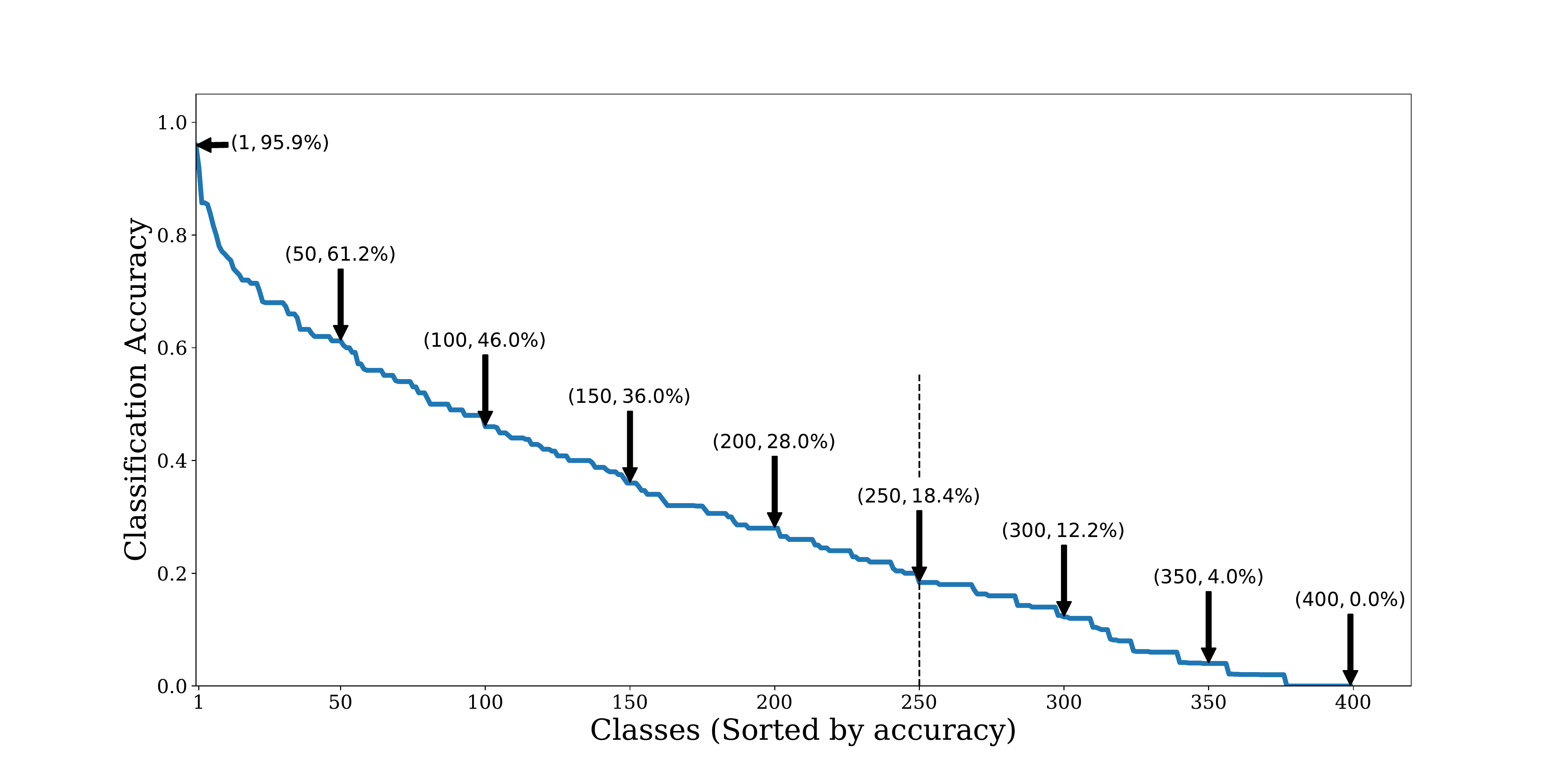} \vspace{0.0cm}
\end{tabular}
\caption{{\bf Kinetics Classes by Classification Accuracy:} Presented are the sorted \emph{top-1} accuracy values for \emph{Kinetics-400} classes. Each tuple denotes class ranking and training classification accuracy, as achieved by the classification method proposed by Yan~\etal~\cite{Yan2018SpatialTG}. The dashed line shows the cut-off accuracy used for selecting classes to be included in \emph{Kinetics-250}. }

\label{figure:supp_acc_by_class}
\end{figure*}

Figure~\ref{figure:supp_acc_by_class} provides the \emph{top-1} training classification accuracy achieved by Yan~\etal~\cite{Yan2018SpatialTG} for each class in \emph{Kinetics-400} in descending order. It is used to show our cutoff point for choosing the \emph{Kinetics-250 classes}. 





\begin{table*}[!t]
\centering
{\small \begin{tabular}{l ccccc| ccccc} 
\toprule

 &  \multicolumn{5}{c}{\textbf{Few vs.\ Many}} & \multicolumn{5}{c}{\textbf{Many vs.\ Few}} \\
\cmidrule(lr){2-6} \cmidrule(lr){7-11}
\textbf{Method} & \textit{Rec. Loss} & \textit{OC-SVM} & \textit{FFP~\cite{Liu_2018}} & \textit{Ours} & \textit{Sup.} & \textit{Rec. Loss} & \textit{OC-SVM} & \textit{FFP~\cite{Liu_2018}} & \textit{Ours} & \textit{Sup.} \\
\midrule	

Arms        & 0.58 & 0.77 & 0.67 & \underline{\textbf{0.86}} & 0.69 & 0.31 & 0.67 & 0.67 & \textbf{0.73} & \underline{0.97} \\
Brushing    & 0.41 & 0.64 & 0.69 & \textbf{0.74} & \underline{0.86} & 0.66 & 0.58 & 0.70 & \textbf{0.73} & \underline{0.97} \\
Dressing    & 0.60 & 0.68 & 0.62 & \textbf{0.80} & \underline{0.87} & 0.61 & 0.74 & 0.63 & \textbf{0.80} & \underline{0.86} \\
Dropping    & 0.42 & 0.71 & 0.62 & \underline{\textbf{0.89}} & 0.87 & 0.47 & 0.68 & 0.61 & \textbf{0.79} & \underline{0.91} \\
Glasses     & 0.49 & 0.77 & 0.51 & \underline{\textbf{0.86}} & 0.82 & 0.41 & 0.66 & 0.55 & \textbf{0.76} & \underline{0.94} \\
Handshaking & 0.87 & 0.51 & 0.70 & \underline{\textbf{0.99}} & 0.90 & \underline{\textbf{0.87}} & 0.85 & 0.72 & 0.71 & 0.71 \\
Office      & 0.45 & 0.56 & 0.71 & \textbf{0.73} & \underline{0.84} & 0.43 & 0.57 & 0.62 & \textbf{0.69} & \underline{0.91} \\
Fighting    & 0.81 & 0.76 & 0.62 & \underline{\textbf{0.99}} & 0.78 & 0.77 & 0.84 & 0.61 & \underline{\textbf{0.99}} & 0.88 \\
Touching    & 0.40 & 0.68 & 0.60 & \underline{\textbf{0.78}} & 0.72 & 0.40 & 0.64 & 0.55 & \textbf{0.66} & \underline{0.98} \\
Waving      & 0.37 & 0.62 & 0.65 & \textbf{0.71} & \underline{0.90} & 0.38 & 0.59 & 0.65 & \textbf{0.74} & \underline{0.89} \\ \midrule
Average     & 0.54 & 0.67 & 0.64 & \underline{\textbf{0.84}} & 0.83 & 0.54 & 0.69 & 0.63 & \textbf{0.76} & \underline{0.90} \\ \midrule

Random 1    & 0.38 & 0.65 & 0.64 & \textbf{0.76} & \underline{0.85} & 0.51 & 0.56 & \textbf{0.65} & 0.61 & \underline{0.95} \\
Random 2    & 0.50 & 0.56 & 0.54 & \textbf{0.72} & \underline{0.79} & 0.50 & 0.58 & 0.51 & \textbf{0.84} & \underline{0.89} \\
Random 3    & 0.64 & 0.54 & 0.54 & \textbf{0.77} & \underline{0.93} & 0.64 & 0.64 & 0.51 & \underline{\textbf{0.84}} & 0.61 \\
Random 4    & 0.38 & 0.66 & 0.73 & \underline{\textbf{0.79}} & 0.74 & 0.43 & 0.59 & \textbf{0.71} & 0.62 & \underline{0.96} \\
Random 5    & 0.53 & 0.53 & 0.50 & \textbf{0.58} & \underline{0.83} & 0.51 & 0.53 & 0.52 & \textbf{0.59} & \underline{0.78} \\
Random 6    & 0.41 & 0.64 & 0.54 & \textbf{0.80} & \underline{0.89} & 0.45 & 0.63 & 0.54 & \textbf{0.75} & \underline{0.94} \\
Random 7    & 0.44 & 0.53 & 0.51 & \textbf{0.66} & \underline{0.87} & 0.46 & 0.52 & 0.51 & \textbf{0.63} & \underline{0.74} \\
Random 8    & 0.65 & 0.54 & 0.57 & \underline{\textbf{0.85}} & 0.82 & 0.63 & 0.72 & 0.56 & \underline{\textbf{0.89}} & 0.78 \\
Random 9    & 0.45 & \textbf{0.69} & 0.52 & 0.62 & \underline{0.86} & 0.48 & \textbf{0.62} & 0.52 & 0.61 & \underline{0.81} \\
Random 10   & 0.61 & 0.64 & 0.54 & \textbf{0.75} & \underline{0.93} & 0.64 & 0.54 & 0.55 & \underline{\textbf{0.74}} & 0.67 \\ \midrule 
Average    & 0.50 & 0.60 & 0.57 & \textbf{0.73} & \underline{0.86} & 0.53 & 0.60 & 0.56 & \textbf{0.69} & \underline{0.82} \\

\bottomrule
\end{tabular}}
\vspace{0.08cm}		
\caption{ \textbf{Coarse Grained Experiment Results - \textit{NTU-RGB+D}:}  Values represent area under the curve (AUC). In bold are the results of the best performing \emph{unsupervised} method. Underlined is the best method of all. \emph{``Sup.''} denotes the supervised baseline.
\emph{``FFP''} denotes the Future frame prediction method by Liu~\etal~\cite{Liu_2018}.
} 
\label{tables:supp-results-ntu}
\vspace{-0.18cm}		
\end{table*}

\begin{table*}[!t]
\centering
{\small \begin{tabular}{l cccccc| cccccc} 
\toprule

 &  \multicolumn{6}{c}{\textbf{Few vs.\ Many}} & \multicolumn{6}{c}{\textbf{Many vs.\ Few}} \\
\cmidrule(lr){2-7} \cmidrule(lr){8-13}
\textbf{Method} & \textit{Rec.} & \textit{OC-SVM} & \textit{FFP~\cite{Liu_2018}} & \textit{TBAD~\cite{Morais_2019_CVPR}} & \textit{Ours} & \textit{Sup.} & \textit{Rec.} & \textit{OC-SVM} & \textit{FFP~\cite{Liu_2018}} & \textit{TBAD~\cite{Morais_2019_CVPR}} & \textit{Ours} & \textit{Sup.} \\
\midrule	

Batting   & 0.40 & 0.46 & 0.58 & 0.64 & \underline{\textbf{0.86}} & 0.76 & 0.55 & 0.46 & 0.57 & 0.64 & \textbf{0.77} & \underline{0.90} \\
Cycling   & 0.41 & 0.56 & 0.61 & 0.59 & \underline{\textbf{0.80}} & 0.63 & 0.62 & 0.54 & 0.64 & 0.54 & \textbf{0.68} & \underline{0.81} \\
Dancing   & 0.30 & 0.63 & 0.53 & 0.68 & \underline{\textbf{0.87}} & 0.73 & 0.84 & 0.37 & 0.54 & 0.57 & \underline{\textbf{0.97}} & 0.87 \\
Gym       & 0.56 & 0.54 & 0.57 & 0.59 & \textbf{0.74} & \underline{0.83} & 0.50 & 0.61 & 0.54 & 0.60 & \underline{\textbf{0.74}} & 0.58 \\
Jumping   & 0.44 & 0.42 & 0.61 & 0.53 & \underline{\textbf{0.70}} & 0.52 & 0.65 & 0.49 & 0.59 & 0.52 & \textbf{0.67} & \underline{0.80} \\
Lifters   & \textbf{0.68} & 0.62 & 0.62 & 0.64 & 0.61 & \underline{0.79} & 0.57 & 0.51 & 0.57 & 0.58 & \textbf{0.70} & \underline{0.84} \\
Music     & 0.43 & 0.52 & 0.61 & 0.60 & \textbf{0.82} & \underline{0.90} & 0.57 & 0.50 & 0.59 & 0.64 & \underline{\textbf{0.62}} & \underline{0.62} \\
Riding    & 0.49 & \textbf{0.61} & 0.60 & 0.53 & 0.56 & \underline{0.66} & 0.65 & 0.52 & 0.61 & 0.55 & \textbf{0.76} & \underline{0.88} \\
Skiing    & 0.42 & 0.45 & \underline{\textbf{0.71}} & 0.54 & 0.68 & 0.62 & 0.54 & 0.51 & \textbf{0.63} & 0.51 & 0.59 & \underline{0.90} \\
Throwing  & 0.41 & 0.58 & 0.51 & 0.6 & \textbf{0.68} & \underline{0.70} & 0.62 & 0.46 & 0.53  & 0.65 & \textbf{0.90} & \underline{0.95} \\ \midrule
Average   & 0.46 & 0.56 & 0.60 & 0.59 & \underline{\textbf{0.73}} & 0.71 & 0.61 & 0.47 & 0.58 & 0.58 & \textbf{0.74} & \underline{0.82} \\ \midrule

Random 1  & 0.48 & \textbf{0.55} & \textbf{0.55} & 0.53 & 0.53 &    \underline{0.81} & 0.39 & 0.61 & 0.58 & 0.54 & \underline{\textbf{0.63}} & 0.58 \\ 
Random 2  & 0.42 & 0.55 & 0.56 & 0.65 & \underline{\textbf{0.71}} &            0.69 & 0.49 & 0.39 & 0.56  & \textbf{0.59} & 0.57 & \underline{0.87} \\ 
Random 3  & 0.49 & 0.54 & 0.55 & 0.56 & \textbf{0.70} &             \underline{0.75} & \underline{\textbf{0.57}} & 0.49 & 0.44 & 0.55 & 0.55 & 0.53 \\ 
Random 4  & 0.49 & 0.48 & 0.48 & 0.56 & \underline{\textbf{0.56}} & \underline{0.56} & 0.53 & 0.43 & 0.52 & 0.51 & \underline{\textbf{0.65}} & 0.56 \\ 
Random 5  & 0.41 & 0.60 & 0.61 & 0.61 & \textbf{0.71} &             \underline{0.76} & \underline{\textbf{0.66}} & 0.41 & 0.57 & 0.52 & 0.62 & 0.56 \\
Random 6  & 0.46 & 0.66 & 0.54 & 0.54 & \underline{\textbf{0.94}}            & 0.90 & 0.57 & 0.56 & 0.49  & 0.62 & \underline{\textbf{0.87}} & 0.56 \\ 
Random 7  & 0.38 & 0.46 & \textbf{0.59} & 0.54 & 0.57 &             \underline{0.67} & 0.48 & 0.45 & \underline{\textbf{0.59}} & 0.54 & 0.54 & 0.54 \\
Random 8  & 0.37 & 0.53 & 0.56 & 0.56 & \textbf{0.63} &             \underline{0.88} & 0.50 & \textbf{0.69} & 0.56 & 0.56 & 0.65 & \underline{0.77} \\
Random 9  & 0.40 & 0.56 & 0.52 & 0.64 & \textbf{0.59} &             \underline{0.80} & \textbf{0.59} & 0.49 & 0.54 & 0.57 & 0.55 & \underline{0.76} \\
Random 10 & 0.52 & \textbf{0.59} & 0.53 & \textbf{0.59} & 0.52 &    \underline{0.85} & 0.54 & \underline{\textbf{0.60}} & \underline{\textbf{0.60}} & 0.57 & 0.53 & 0.52 \\ \midrule
Average   & 0.45 & 0.56 & 0.55 & 0.57 & \textbf{0.65} & \underline{0.77} & 0.51 & 0.52 & 0.55 & 0.56 & \textbf{0.62} & \underline{0.63} \\
\bottomrule
\end{tabular}}
\vspace{0.08cm}		
\caption{ \textbf{Coarse Grained Experiment Results - \textit{Kinetics-250}:}  Values represent area under the curve (AUC). In bold are the results of the best performing \emph{unsupervised} method. Underlined is the best method of all. \emph{``Sup.''} denotes the supervised baseline.
\emph{``FFP''} denotes the Future frame prediction method by Liu~\etal~\cite{Liu_2018}.
\emph{``TBAD''} denotes the Trajectory based anomaly detection method by Morais~\etal~\cite{Morais_2019_CVPR}.
}
\label{tables:supp-results-kinetics}
\end{table*}

\begin{table*}[h]
\centering
{\small \begin{tabular}{c l} 
\toprule

\multicolumn{2}{c}{\textbf{NTU-RGB+D}} \\ \midrule

Arms        & Pointing to something with finger (31), Salute (38), Put the palms together (39), Cross hands in front (say stop) (40) \\ \midrule
Brushing    & Drink water (1), Brushing teeth (3), Brushing hair (4) \\ \midrule
Dressing    & Wear jacket (14), Take off jacket (15), Wear a shoe (16), Take off a shoe (17) \\ \midrule
Dropping    & Drop (5), Pickup (6), Sitting down (8), Standing up (from sitting position) (9) \\ \midrule
Glasses     & Wear on glasses (18), Take off glasses (19), Put on a hat/cap (20), Take off a hat/cap (21) \\ \midrule
Handshaking & Hugging other person (55), Giving something to other person (56), Touch other person's pocket (57), Handshaking (58) \\ \midrule
Office      & Make a phone call/answer phone (28), Playing with phone/tablet (29), Typing on a keyboard (30), \\ & Check time (from watch) (33) \\ \midrule
Fighting    & Punching/slapping other person (50), Kicking other person (51), Pushing other person (52), Pat on back of other person (53) \\ \midrule
Touching    & Touch head (headache) (44), Touch chest (stomachache/heart pain) (45), Touch back (backache) (46), \\ & Touch neck (neckache) (47) \\ \midrule
Waving      & Clapping (10), Hand waving (23), Pointing to something with finger (31), Salute (38) \\ \midrule
Random 1    & Brushing teeth (3), Pointing to something with finger (31), Nod head/bow (35), Salute (38) \\ \midrule
Random 2    & Walking apart from each other (0), Throw (7), Wear on glasses (18), Hugging other person (55) \\ \midrule
Random 3    & Brushing teeth (3), Tear up paper (13), Wear jacket (14), Staggering (42) \\ \midrule
Random 4    & Eat meal/snack (2), Writing (12), Taking a selfie (32), Falling (43) \\ \midrule
Random 5    & Playing with phone/tablet (29), Check time (from watch) (33), Rub two hands together (34), Pushing other person (52) \\ \midrule
Random 6    & Eat meal/snack (2), Take off glasses (19), Take off a hat/cap (21), Kicking something (24) \\ \midrule
Random 7    & Drop (5), Tear up paper (13), Wear on glasses (18), Put the palms together (39) \\ \midrule
Random 8    & Falling (43), Kicking other person (51), Point finger at the other person (54), Point finger at the other person (54) \\ \midrule
Random 9    & Wear on glasses (18), Rub two hands together (34), Falling (43), Punching/slapping other person (50) \\ \midrule
Random 10   & Throw (7), Clapping (10), Use a fan (with hand or paper)/feeling warm (49), Giving something to other person (56) \\ 

\bottomrule
\end{tabular}}
\vspace{0.08cm}		
\caption{\small{\textbf{Complete List of Splits - NTU-RGB+D:} The splits used for evaluation for {\it NTU-RGB+D} dataset. Numbers are the numeric class labels. Often split names carry no significance and were chosen to be one of the split classes. }}
\vspace{-0.34cm}
\label{table:supp_splits_ntu}
\end{table*}	

\begin{table*}[ht!]
\centering
{\small \begin{tabular}{c l} 
\toprule

\multicolumn{2}{c}{\textbf{Kinetics}} \\
\midrule            
Batting & Golf driving (143), Golf putting (144), Hurling (sport) (162), Playing squash or racquetball (246), Playing tennis (247) \\ \midrule
Cycling & Riding a bike (268), Riding mountain bike (272), Riding unicycle (276), Using segway (376) \\ \midrule
Dancing & Belly dancing (19), Capoeira (44), Country line dancing (76), Salsa dancing (284), Tango dancing (349), Zumba (400) \\ \midrule
Gym & Lunge (184), Pull Ups (265), Push Up (261), Situp (306), Squat (331) \\ \midrule
Jumping &  High jump (152), Jumping into pool (173), Long jump (183), Triple jump (368) \\ \midrule
Lifters &  Bench pressing (20), Clean and jerk (60), Deadlifting (89), Front raises (135), Snatch weight lifting (319) \\ \midrule
Music & Playing accordion (218), Playing cello (224), Playing clarinet (226), Playing drums (231), Playing guitar (233), \\ & Playing harp (235)  \\ \midrule
Riding & Lunge (184), Pull Ups (256), Push Up (261), Situp (306), Squat (331) \\ \midrule
Skiing & Roller skating (281), Skateboarding (307), Skiing slalom (311), Tobogganing (361) \\ \midrule
Throwing & Hammer throw (149), Javelin throw (167), Passing american football (in game) (209), Shot put (299), Throwing axe (357), \\ & Throwing discus (359) \\ \midrule
Random 1 & Climbing tree (69), Juggling fire (171), Marching (193), Shaking head (290), Using segway (376) \\ \midrule
Random 2 & Drop kicking (106), Golf chipping (142), Pole vault (254), Riding scooter (275), Ski jumping (308) \\ \midrule
Random 3 & Bench pressing (20), Hammer throw (149), Playing didgeridoo (230), Sign language interpreting (304), \\ & Wrapping present (395) \\ \midrule
Random 4 & Cleaning floor (61), Ice fishing (164), Using segway (376), Waxing chest (388) \\ \midrule
Random 5 & Barbequing (15), Golf chipping (142), Kissing (177), Lunge (184) \\ \midrule
Random 6 & Arm wrestling (7), Crawling baby (78), Presenting weather forecast (255), Surfing crowd (337) \\ \midrule
Random 7 & Bobsledding (29), Canoeing or kayaking (43), Dribbling basketball (100), Playing ice hockey (236)\\ \midrule
Random 8 & Playing basketball (221), Playing tennis (247), Squat (331) \\ \midrule
Random 9 & Golf putting (144), Juggling fire (171), Walking the dog (379) \\ \midrule
Random 10 & Jumping into pool (173), Krumping (180), Presenting weather forecast (255) \\ \bottomrule
\end{tabular}}
\vspace{0.08cm}		
\caption{\small{\textbf{Complete List of Splis - Kinetics-250:} The splits used for evaluation for {\it Kinetics-250} dataset. Numbers are the numeric class labels. Often split names carry no significance and were chosen to be one of the split classes.}}
\vspace{-0.4cm}
\label{table:supp_splits_kinetics}
\end{table*}	

\clearpage
\clearpage
\section{Kinetics-250 Class List} 
\label{supp_kin250}

{\small
\begin{enumerate}
\setlength{\itemsep}{1pt}
\setlength{\parskip}{0pt}
\setlength{\parsep}{0pt}
\raggedbottom
\item Abseiling (1)   
\item Air drumming (2)
\item Archery (6)
\item Arm wrestling (7)
\item Arranging flowers (8)
\item Assembling computer (9)
\item Auctioning (10)
\item Barbequing (15)
\item Bartending (16)
\item Beatboxing (17)
\item Belly dancing (19)
\item Bench pressing (20)
\item Bending back (21)
\item Biking through snow (23)
\item Blasting sand (24)
\item Blowing glass (25)
\item Blowing out candles (28)
\item Bobsledding (29) 
\item Bookbinding (30)
\item Bouncing on trampoline (31)
\item Bowling (32)
\item Braiding hair (33)
\item Breakdancing (35)
\item Building cabinet (39)
\item Building shed (40)
\item Bungee jumping (41)
\item Busking (42)
\item Canoeing or kayaking (43)
\item Capoeira (44)
\item Carrying baby (45)
\item Cartwheeling (46)
\item Catching or throwing softball (51)
\item Celebrating (52)
\item Cheerleading (56)
\item Chopping wood (57)
\item Clapping (58)
\item Clean and jerk (60)
\item Cleaning floor (61)
\item Climbing a rope (67)
\item Climbing tree (69)
\item Contact juggling (70)
\item Cooking chicken (71)
\item Country line dancing (76)
\item Cracking neck (77)
\item Crawling baby (78)
\item Curling hair (81)
\item Dancing ballet (85)
\item Dancing charleston (86)
\item Dancing gangnam style (87)
\item Dancing macarena (88)
\item Deadlifting (89)
\item Dining (92)
\item Disc golfing (93)
\item Diving cliff (94)
\item Doing aerobics (96)
\item Doing nails (98)
\item Dribbling basketball (100)
\item Driving car (104)
\item Driving tractor (105)
\item Drop kicking (106)
\item Dunking basketball (108)
\item Dying hair (109)
\item Eating burger (110)
\item Eating spaghetti (117)
\item Exercising arm (120)
\item Extinguishing fire (122)
\item Feeding birds (124)
\item Feeding fish (125)
\item Feeding goats (126)
\item Filling eyebrows (127)
\item Finger snapping (128)
\item Flying kite (131)
\item Folding clothes (132)
\item Front raises (135)
\item Frying vegetables (136)
\item Gargling (138)
\item Giving or receiving award (141)
\item Golf chipping (142)
\item Golf driving (143)
\item Golf putting (144)
\item Grooming horse (147)
\item Gymnastics tumbling (148)
\item Hammer throw (149)
\item Headbanging (150)
\item High jump (152)
\item Hitting baseball (154)
\item Hockey stop (155)
\item Hopscotch (157)
\item Hula hooping (160)
\item Hurdling (161)
\item Hurling (sport) (162)
\item Ice climbing (163)
\item Ice fishing (164)
\item Ice skating (165)
\item Ironing (166)
\item Javelin throw (167)
\item Jetskiing (168)
\item Jogging (169)
\item Juggling balls (170)
\item Juggling fire (171)
\item Juggling soccer ball (172)
\item Jumping into pool (173)
\item Jumpstyle dancing (174)
\item Kicking field goal (175)
\item Kicking soccer ball (176)
\item Kissing (177)
\item Knitting (179)
\item Krumping (180)
\item Laughing (181)
\item Long jump (183)
\item Lunge (184)
\item Making bed (187)
\item Making snowman (190)
\item Marching (193)
\item Massaging back (194)
\item Milking cow (198)
\item Motorcycling (200)
\item Mowing lawn (202)
\item News anchoring (203)
\item Parkour (208)
\item Passing american football (in game) (209)
\item Passing american football (not in game)(210)
\item Picking fruit (215)
\item Playing accordion (218)
\item Playing badminton (219)
\item Playing bagpipes (220)
\item Playing basketball (221)
\item Playing bass guitar (222)
\item Playing cello (224)
\item Playing chess (225)
\item Playing clarinet (226)
\item Playing cricket (228)
\item Playing didgeridoo (230)
\item Playing drums (231)
\item Playing flute (232)
\item Playing guitar (233)
\item Playing harmonica (234)
\item Playing harp (235)
\item Playing ice hockey (236)
\item Playing kickball (238)
\item Playing organ (240)
\item Playing paintball (241)
\item Playing piano (242)
\item Playing poker (243)
\item Playing recorder (244)
\item Playing saxophone (245)
\item Playing squash or racquetball (246)
\item Playing tennis (247)
\item Playing trombone (248)
\item Playing trumpet (249)
\item Playing ukulele (250)
\item Playing violin (251)
\item Playing volleyball (252)
\item Playing xylophone (253)
\item Pole vault (254)
\item Presenting weather forecast (255)
\item Pull ups (256)
\item Pumping fist (257)
\item Punching bag (259)
\item Punching person (boxing) (260)
\item Push up (261)
\item Pushing car (262)
\item Pushing cart (263)
\item Reading book (265)
\item Riding a bike (268)
\item Riding camel (269)
\item Riding elephant (270)
\item Riding mechanical bull (271)
\item Riding mountain bike (272)
\item Riding or walking with horse (274)
\item Riding scooter (275)
\item Riding unicycle (276)
\item Robot dancing (278)
\item Rock climbing (279)
\item Rock scissors paper (280)
\item Roller skating (281)
\item Running on treadmill (282)
\item Sailing (283)
\item Salsa dancing (284)
\item Sanding floor (285)
\item Scrambling eggs (286)
\item Scuba diving (287)
\item Shaking head (290)
\item Shaving head (293)
\item Shearing sheep (295)
\item Shooting basketball (297)
\item Shot put (299)
\item Shoveling snow (300)
\item Shuffling cards (302)
\item Side kick (303)
\item Sign language interpreting (304)
\item Singing (305)
\item Situp (306)
\item Skateboarding (307)
\item Ski jumping (308)
\item Skiing (not slalom or crosscountry) (30
\item Skiing crosscountry (310)
\item Skiing slalom (311)
\item Skipping rope (312))
\item Skydiving (313)
\item Slacklining (314)
\item Sled dog racing (316)
\item Smoking hookah (318)
\item Snatch weight lifting (319)
\item Snorkeling (322)
\item Snowkiting (324)
\item Spinning poi (327)
\item Springboard diving (330)
\item Squat (331)
\item Stomping grapes (333)
\item Stretching arm (334)
\item Stretching leg (335)
\item Strumming guitar (336)
\item Surfing crowd (337)
\item Surfing water (338)
\item Sweeping floor (339)
\item Swimming backstroke (340)
\item Swimming breast stroke (341)
\item Swimming butterfly stroke (342)
\item Swinging legs (344)
\item Tai chi (347)
\item Tango dancing (349)
\item Tap dancing (350)
\item Tapping guitar (351)
\item Tapping pen (352)
\item Tasting beer (353)
\item Testifying (355)
\item Throwing axe (357)
\item Throwing discus (359)
\item Tickling (360)
\item Tobogganing (361)
\item Training dog (364)
\item Trapezing (365)
\item Trimming or shaving beard (366)
\item Triple jump (368)
\item Tying tie (371)
\item Using segway (376)
\item Vault (377)
\item Waiting in line (378)
\item Walking the dog (379)
\item Washing feet (381)
\item Water skiing (384)
\item Waxing chest (388)
\item Waxing eyebrows (389)
\item Welding (392)
\item Windsurfing (394)
\item Wrapping present (395)
\item Wrestling (396)
\item Yoga (399)
\item Zumba (400)
\end{enumerate}
}

\end{document}